\documentclass[sn-mathphys-num]{sn-jnl}

\usepackage{graphicx}%
\usepackage{multirow}%
\usepackage{amsmath,amssymb,amsfonts}%
\usepackage{amsthm}%
\usepackage{mathrsfs}%
\usepackage[title]{appendix}%
\usepackage{xcolor}%
\usepackage{textcomp}%
\usepackage{manyfoot}%
\usepackage{booktabs}%
\usepackage{algorithm}%
\usepackage{algorithmicx}%
\usepackage{algpseudocode}%
\usepackage{listings}%
\usepackage{caption}
\usepackage{subcaption}
\usepackage{semantic}
\usepackage{bm}
\usepackage{tikz}
\usepackage{tikz-3dplot}

\usetikzlibrary{calc}
\usetikzlibrary{backgrounds}
\usetikzlibrary{shapes,arrows,positioning}
\usetikzlibrary{patterns}
\tdplotsetmaincoords{60}{125}
\tdplotsetrotatedcoords{0}{0}{0}
\usetikzlibrary{decorations.markings}
\tikzstyle arrowstyle=[scale=1.4]
\tikzstyle directed=[postaction={decorate,decoration={markings,
    mark=at position .65 with {\arrow[arrowstyle]{stealth}}}}]

\usepackage{pgfplots}
\usepackage{units}
\usepackage{physics}
\newlength\fwidth
\DeclareMathOperator{\atantwo}{atan2}
\DeclareUnicodeCharacter{2217}{*}

\begin{document}

\title[Article Title]{DTAA: A Detect, Track and Avoid Architecture for navigation in spaces with Multiple Velocity Objects} 


\author*[1]{\fnm{Samuel} \sur{Nordstr{\"o}m}}\email{samuel.nordstrom@ltu.se}

\author[1]{\fnm{Bj{\"o}rn} \sur{Lindquist}}\email{bjorn.lindqvist@ltu.se}

\author[1]{\fnm{George} \sur{Nikolakopoulos}}\email{george.nikolakopoulos@ltu.se}

\affil*[1]{\orgdiv{Robotics \& AI Team in department of Computer, Electrical and Space}, \orgname{Lule\r{a} University of Tehnology}, \orgaddress{\street{Street}, \city{Lule\r{a}}, \postcode{100190}, \state{Norrbotten}, \country{Sweden}}}


\abstract{
Proactive collision avoidance measures are imperative in environments where humans and robots coexist. Moreover, the introduction of high-quality legged robots into workplaces highlighted the crucial role of a robust, fully autonomous safety solution for robots to be viable in shared spaces or in co-existence with humans. This article establishes for the first time ever an innovative Detect-Track-and-Avoid Architecture (DTAA) to enhance safety and overall mission performance. The proposed novel architecture has the merit ot integrating object detection using YOLOv8, utilizing Ultralytics embedded object tracking, and state estimation of tracked objects through Kalman filters. Moreover, a novel heuristic clustering is employed to facilitate active avoidance of multiple closely positioned objects with similar velocities, creating sets of unsafe spaces for the Nonlinear Model Predictive Controller (NMPC) to navigate around.
The NMPC identifies the most hazardous unsafe space, considering not only their current positions but also their predicted future locations. The NMPC calculates maneuvers to guide the robot along a path planned by D$^{*}_{+}$ towards its intended destination, while maintaining a safe distance to all identified obstacles. The efficacy of the novelty suggested DTAA framework is being validated by Real-life experiments featuring a Boston Dynamics Spot robot that demonstrates the robot's capability to consistently maintain a safe distance from humans in dynamic subterranean, urban indoor, and outdoor environments. 
}




\maketitle

\section{Introduction}

In the dynamic landscapes of construction sites or mines, where human workers, vehicles, and other robots coexist, mitigating the risk of accidents becomes of paramount importance. Ensuring safety as become a gatekeeper for any robot or machine to enter environments with mixed traffic.

In recent times, there has been a notable surge in the integration of local navigation and the avoidance of moving obstacles, attracting considerable attention as evidenced by various methods~\cite{mohanan2018survey}, such as dynamic windows~\cite{xinyi2019dynamic}, an approach that has been synergistically combined with reinforcement learning~\cite{9561462}, while other research approaches have used probability calculations~\cite{gopalakrishnan2017prvo} to avoid obstacles. In the specific case where the moving obstacle in consideration is a human, it is referred as ``social navigation'' an area that has recently received a lot of attention~\cite{singamaneni2024survey}. Humans are a special kind of moving obstacle where their behavior is affected by what is around them, and they can feel uncomfortable with robots in close proximity to them, resulting in a desire to have the robots behave in a way that makes humans comfortable. Because it is hard to put numbers on humans' comfort levels this article focuses on the measurable safety distances as a clear indicator for the evaluation of the proposed system. 

Our innovative framework is based on the Non-linear Model Predictive Control (NMPC)~\cite{lindqvist2020dynamic,5984510}, offering a fundamental framework that enables not only reactive but also proactive avoidance. This proactive capability stems from the NMPC's ability to predict future collisions, thereby initiating avoidance maneuvers well in advance. Model Predictive Control (MPC), owing to its predictive nature and direct connection to the robot's dynamics, emerges as an ideal candidate for explicit consideration of both cooperative~\cite{lindqvist2021scalable} and non-cooperative~\cite{kamel2017robust} moving obstacles during maneuvering.
Despite the existence of a plethora of contemporary methods addressing MPC problems with considerations for model and obstacle uncertainty, such as Chance Constraints~\cite{zhang2020trajectory} and Robust Scenario MPC~\cite{batkovic2020robust}, the literature lacks substantial exploration of the essential link between onboard perception systems and MPC for velocity obstacles. This connection and its subsequent experimental verification are crucial for real-world implementations in unfamiliar environments.
Apart from the previous work in~\cite{karlsson2022ensuring}, that have addressed onboard tracking/estimation and presented experimental results in a real robot-human safety scenario in~\cite{lin2020robust}, other approaches include the one in~\cite{9561326} utilizing an MPC to do low-level collision-free motion planning for quadruped-legged robots with a lidar to detect the objects in the environment.
\begin{figure}
    \centering
    \resizebox{1.0\columnwidth}{!}{%
\begin{tikzpicture}
    \node at (0,0) {\includegraphics[width=0.7\columnwidth]{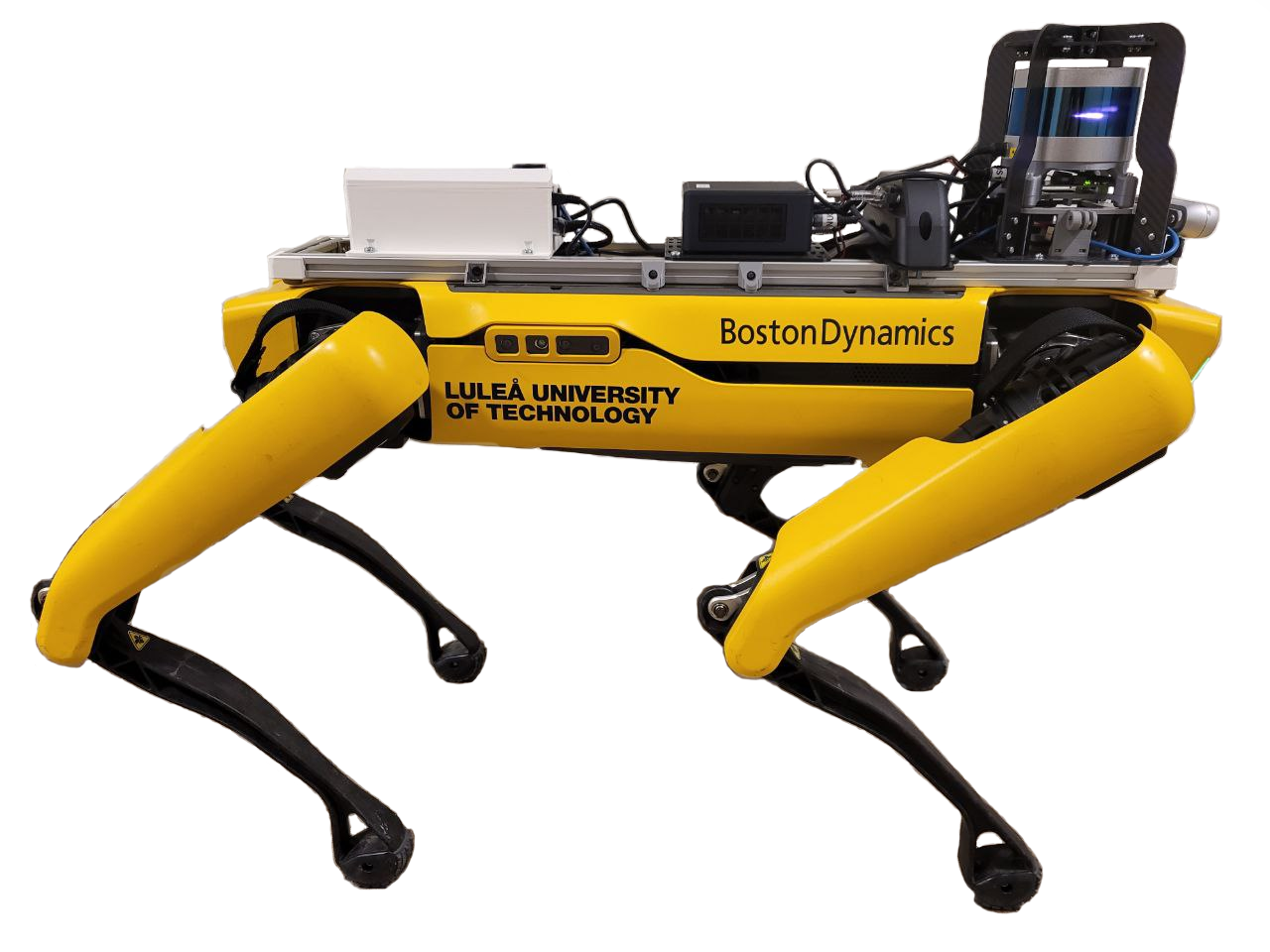}}; 

    \node[] at (3.2, 3.7) (vlp) {Velodyne};
    \node[align=center] at (0.8, 3.0) (orin) {Jetson\\AGX Orin};
    \node[align=center] at (5.0, 1.9) (rs) {Camera\\IMU};
     
    \draw[-stealth] (vlp) -> (3.2, 2.8);
    \draw[-stealth] (orin) -> (0.8, 2.15);
    \draw[-stealth] (rs) -> (4.1, 1.9);
\end{tikzpicture}
}
    \caption{Spot, the robot used for experimentation with the added sensors on top.}
    \label{fig:spot}
\end{figure}
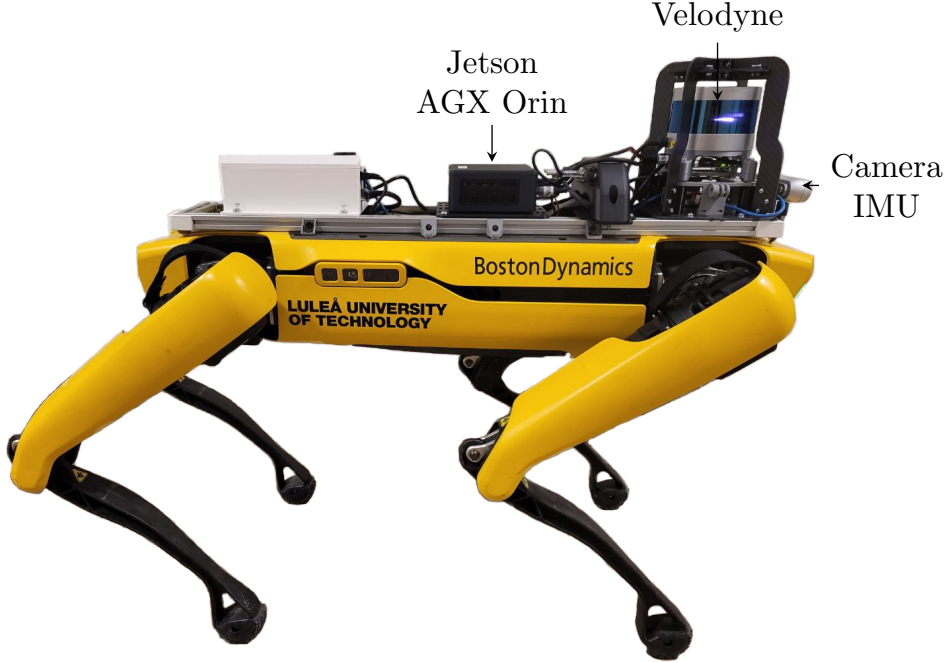

In our previous work, where only one velocity object tracked at any time, leading to some severe weaknesses, since as soon as a second obstacle entered the frame it failed to correctly track the obstacle, which could lead to dangerous errors in avoidance planning. As in the previous work, we utilize a Convolutional Neural Network~\cite{cai2016unified} (CNN) to detect and classify objects of interest (humans, other robots, vehicles) in a video feed from onboard sensors. The specific state-of-the-art CNN detector~\cite{terven2023comprehensive}, utilized in this work, is YOLOv8~\cite{jocher2023yolo}, 
and it is combined with multiple Kalman filters~\cite{welch1995introduction} to estimate the obstacle's positions and velocity's. 

In general, object tracking has been extensively studied in various setups and constraints as in~\cite{robin2016multi}. Particularly in robotics, effective temporary tracking involves recognizing color patterns~\cite{nummiaro2003color, nummiaro2002object}, while approaches incorporating predicted motion models for improved accuracy have appeared in~\cite{stauffer2000learning}. Some solutions have fused tracking and detection mechanisms for enhanced results, as demonstrated in~\cite{zhou2020tracking}. However, this approach increased complexity and reduced the dynamic adaptability. It should be also noted that tracking approaches vary based on end-user cases, with applications in robotics highlighted in~\cite{saha2022efficient},~\cite{eppenberger2020leveraging},~\cite{wang2018robust}. A persistent challenge in robotics is the reliable tracking of detected objects, during fast radial movements, induced by the robot. To address this, trajectory tracking, as proposed in~\cite{antonini2006counting}, provides a robust solution for accurate tracking, while it has also utilized the tracking feature embedded in YOLOv8\cite{jocher2023yolo}.


Furthermore, the field of clustering is a well-explored research domain, as comprehensively reviewed in~\cite{ezugwu2022comprehensive}, and it has found diverse applications within robotics. Prior research has employed clustering techniques to group buildings, streamlining path planning processes~\cite{li2020obstacle}, and to categorize lidar points effectively~\cite{gao2021dynamic}. In the context of obstacle avoidance, clustering has been utilized, as seen in~\cite{park2020stereo}, primarily focusing on static obstacles and clustering points within a point cloud to identify obstacles. Additionally, the utilization of ellipses to unite clusters for avoidance has been explored in the literature as in~\cite{lee2017velocity}, where it was applied to point clouds for effective grouping. 

\subsection{Contributions}

This article significantly extends our previous work~\cite{karlsson2022ensuring} by introducing a novel framework for multi-obstacle avoidance. In this case, an obstacle clustering approach, combined with the generation of minimal bounding ellipses is linked to a constrained nonlinear model predictive controller. Our novel framework enables the detection, tracking, prediction, dynamic clustering, unsafe set generation, and finally prioritization of pedestrians in proximity to the robot. The result is a streamlined and compact perception pipeline toward real-world human-robot safety scenarios for legged robots, which is evaluated in multiple lab and field experiments.

\section{Detect, Track, and Avoid Architecture (DTAA)}

The proposed Detect, Track, and Avoid Architecture (DTAA) for obstacle avoidance combines multiple components: YOLOv8 (You Only Lock Once), tracking and state estimation, clustering, and a nonlinear model predictive controller. These components are described later in details, as they form the core of the DTAA architecture. Still, to allow realistic evaluations, the proposed framework is combined with a full autonomy kit, which is described as follows and also depicted in Figure~\ref{fig:artchetecture}.
In the presented approach, we utilize LIO-SAM~\cite{liosam2020shan} for real-time state estimation to generate the robot state $\hat{x}$. To test the architecture in more realistic scenarios, the risk-aware path planer $D^{*}_{+}$~\cite{karlsson2022d+} was used to plan a path $\Psi$ to emulate robotic missions. D$^{*}_{+}$ plans $\Psi$ from the robot's current location to any desired waypoint based on a grid map $\mathbf{M}$. In this mission, we use the traversability map BGK$^{+}$~\cite{bayesian2018shan}, which provides a $\mathbf{M}$ indicating where the robot can and can not walk. To enable the full potential of the NMPC, a splined path $\Psi$ with a step length, corresponding to Spot's traversal distance per time step is utilized. This $\Psi$ is shifted based on the output from an Artificial Potential Fields (APF)'s~\cite{lindqvist2022adaptive} to form the reference ($\psi$) that is given to the NMPC as an extra safety layer. The APF operates directly on raw LiDAR point clouds $\Pi_{pcl}$, where each beam with a distance $<s$, adds a repellent force that is used to shift the original trajectory $\Psi$. 
\input{artchetecture.tex}

A Boston Dynamics Spot robot (see figure~\ref{fig:spot}) was used to test the avoidance architecture that was equipped with an Intel RealSense D455 camera that captures images in the image frame $\mathcal{I}$. This frame has a static transformation to the robot body frame $\mathcal{B}$. The $\mathcal{B}$'s relation to the global frame $\mathcal{G}$ is dynamic and is provided by LIO-SAM, which uses data from the Velodyne VLP-16 3D lidar and the Vectornav vn-100 inertial measurement unit also onboard the robot, while the corresponding frames interconnection is visualized in Figure~\ref{fig:framesD}. Finally, it should be noted that all the computations are executed live on the onboard Jetson AGX Orin computer.

\input{cordinateFrames.tex}

\subsection{Design choices}
\begin{figure}
    \centering
    \setlength\fwidth{0.95\columnwidth}
%
%
\definecolor{mycolor1}{rgb}{0.00000,0.44700,0.74100}%
\begin{tikzpicture}

\begin{axis}[%
width=0.8\fwidth,
height=0.4\fwidth,
at={(0.9\fwidth,0.9\fwidth)},
scale only axis,
xmin=0,
xmax=65,
xlabel style={font=\color{white!15!black}},
xlabel={Frames per second (fps)},
ymin=0.0,
ymax=0.24,
ylabel style={font=\color{white!15!black}},
ylabel={Time (s)},
axis background/.style={fill=white},
title style={font=\bfseries},
title={Computation delay},
axis x line*=bottom,
axis y line*=left,
legend style={at={(0.03,0.97)}, anchor=north west, legend cell align=left, align=left, draw=white!15!black}
]
\addplot [color=mycolor1, line width=2.0pt]
  table[row sep=crcr]{%
5	0.0483175196890104\\
10	0.0476061478686195\\
15	0.0428136430146151\\
20	0.0428136430146151\\
25	0.0456604617834091\\
33	0.0467062724521524\\
38	0.195845134179671\\
42	0.227148414314581\\
45	0.0679309829354541\\
51	0.119220537404571\\
55	0.17565103719565\\
60	0.203034723704447\\
};
\addlegendentry{Avrage delay}

\end{axis}
\end{tikzpicture}%
    \caption{Total computation delay from the time instant that an image is initially received until the NMPC has calculated a control output versus different frame rates for the utilized camera.}
    \label{plot:delay}
\end{figure}
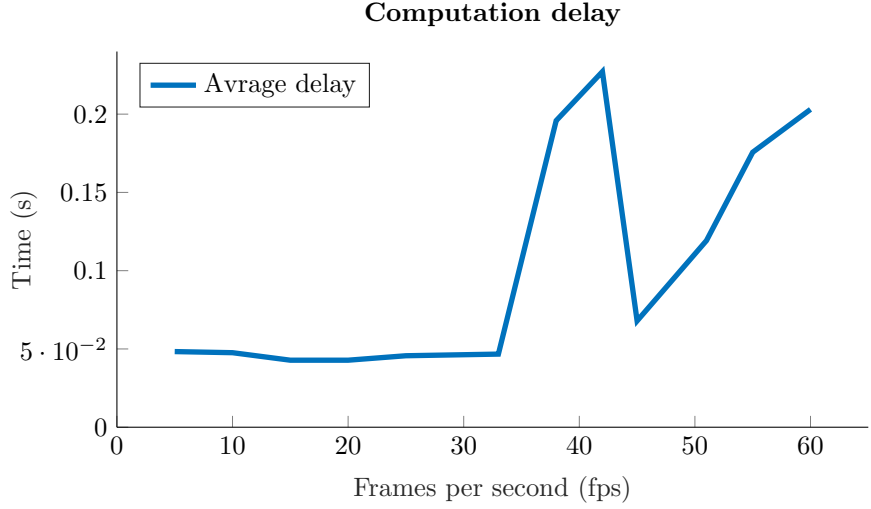

For a successful obstacle avoidance, it is important that obstacles are detected as fast as possible so avoidance maneuvers can be initiated as early as possible, giving the best chance to get out of collision in time. Velocity estimations also benefit from a higher frame rate, given more data points to estimate a more accurate velocity. In this research, the state-of-the-art object detection YOLOv8 algorithm has been used~\cite{terven2023comprehensive}, with the framework provided by the creator~\cite{jocher2023yolo}.
The computer used, Jetson AGX Orin, is capable of object detection on a single image in approximately $\unit[30]{ms}$, counted from the time that the image is received, until the data is extracted and ready to be used for tracking and clustering. A significant part of the time is consumed in loading the image to and from the GPU. To achieve a higher execution frame rate the detection should be implemented in a multi-threaded approach, resulting in a detection being executed 60 times a second but with the same delay of approximately $\unit[30]{ms}$, when testing the detection isolated. However, when the the full system is operational (full autonomy stack) the delays approximately are 10 times higher, mainly due to the extra load on the entire computer, creating a lot more random delays and interrupts. In Figure~\ref{plot:delay}, the switch to multi-threading is visible by the significantly increased delays. In addition, the increased delays are relaxing the need for an increased frame rate and therefore the single-threaded approach has been selected and implemented with a $\unit[30]{hz}$ frame rate.

The ultralytics framework~\cite{terven2023comprehensive} is also capable of tracking the objects it detects over time.
This feature requires a sequential execution of the images and is therefore not possible to run with the multi-threaded approach.
A comparison in tracking precision between our greedy matching implementation (presented in Section~\ref{sec:greed}) and ultralytics is performed with one human in the testing laboratory. During that test, the ultralytics continuously track the human by only losing the tracking when the human went out of field of view. When compared to our greedy matching algorithm, it detected $13$ extra pedestrians due to noise that was possible to filter with the ultralytics tracking. In the evaluation scenario where the pedestrian left the field of view and entered again, the ultralytics tracker has  naturally registered the pedestrian as a new pedestrian/object, while our presented greedy matching, still registered this event as the same pedestrian.
This is because our greedy matching is performing in 3D space and can therefore keep track of objects outside of the field of view for a short time.
Finally, due to the noise rejection ability, the ultralytics tracking has been selected for the experiments. in this case, the tracking outside the field of view is still used for avoidance but if the pedestrian repeats the same movement of out and inside the field of view, it will be registered as a new obstacle instead of continuing identifying/recognising it as the same.

\subsection{Detection of humans}
A convolution neural network (CNN) could be used to detect objects in an RGB image ($I_{rgb}$). This article uses the CNN called You Only Look Once version 8 (YOLOv8)~\cite{jocher2023yolo} to detect humans for avoidance. YOLOv8 is a state-of-the-art detection network that runs fast enough to be used in real-time onboard robots, with a frame rate of $\unit[30]{hz}$. The pre-trained weights used in this article are from~\cite{jocher2023yolo} and were trained on the Microsoft Coco data set~\cite{lin2014microsoft} for the nano model of YOLOv8 (``yolov8n.engine'').
Detected humans position in $I_{rgb}$ are provided in bounding boxes $b = [x^\mathcal{I},y^\mathcal{I},H^\mathcal{I},W^\mathcal{I}]$ ($H=$ height, $W=$ width) measured in pixels as illustrated in figure~\ref{fig:framesD}.
When multiple humans are in the camera's field of view will YOLOv8 create one $b$ for each human resulting in a list $B$.

\subsection{Tracking}
To track and estimate the position and velocity of the detected objects in $B$, it is necessary to transform them from the image to the global frame or $\mathcal{I} \to \mathcal{G}$. The depth to the detected object is obtained from the depth image $I_{depth}$ that is aligned with the color image. The average value from a 3x3 square, taken from the objects' center, is utilized to filter some noise. The object position $B_p$, measured in pixels, is determined in the frame $\mathcal{I}$. To get the position in meters $B_m$, the camera intrinsic matrix $\mathbf{K}$ utilized is:
\begin{align}
    \mathbf{K}^{-1}B_p=B_m
    \label{eq:pm}
\end{align}
To transform $B^{\mathcal{I}}$ to the global frame $B^{\mathcal{G}}$, the following transformation is utilized:
\begin{align}
    B^{\mathcal{I}}  {^{\mathcal{G}}T_{\mathcal{I}}}=B^{\mathcal{G}}
\end{align} In this case, the obstacles $W_{obs}$ and $H_{obs}$ are obtained by giving the edge points of $B$ to~\eqref{eq:pm}.

\subsubsection{Greedy matching} \label{sec:greed}
To pair the newly detected objects $B_{new}$ with the previously detected and tracked objects $B_{obs}$, a greedy matching algorithm was implemented and tested.

Assuming that the movements of the objects are relatively small between frames, is it likely that $B_{new} \approx B_{obs}$. Paring the $B_{new}$ and $B_{obs}$, and by starting with the pair that are the most equal in position, results in a greatly matched $B_{new}$ and $B_{obs}$. If the difference between $B_{new}$ and $B_{obs}$ is larger than a threshold, $B_{new}$ is added as a new object to the tracked objects. Old objects that are not matched with a new object are kept for some frames, in case they reappear due to obstruction or field of view limitation.

\subsubsection{Ultralytics embedded tracker}
With the embedded Ultralytics tracking, the objects are already tracked, so all $B_{new}$ have an ID corresponding to an ID in $B_{old}$, unless it is a new obstacle that has appeared.

\subsubsection{Extended Kalman filter}
The old objects' position $B^{\mathcal{G}}$ is updated with the new one by using an extended Kalman filter that also estimates the objects' velocity to get the objects state $\hat{o}^{\mathcal{G}} = [b^{\mathcal{G}}_{x}, b^{\mathcal{G}}_{y}, v^{\mathcal{G}}_{x}, v^{\mathcal{G}}_{y}]$.
The state is assumed to evolve according to:
\begin{align}
    \hat{o}^{\mathcal{G}}_{k+n+1|k} = \mathbf{F} \hat{o}_{k+n|k}^\mathcal{G} + \mathbf{G}a
\end{align}
where:
\begin{align}
    \mathbf{F} &= \left[\begin{array}{cc}
         1 & \delta t  \\   
         0 & 1
    \end{array}\right] \\
    \mathbf{G} &= \left[\begin{array}{c}
      \frac{\delta t^2}{2}  \\
      \delta t
    \end{array}\right]
\end{align}
and $a$ is the acceleration. This is performed for each individual axis and thus three identical extended Kalman filters run in parallel for each object that all result in a set $O$ of obstacles, where $\forall \hat{o} \in O$.
When testing different frame rates for the detection $\delta t$ is changed to match the frame rate.

\subsubsection{Constant velocity assumtion}
In the context of predicting obstacles, we assume that they move at a constant velocity throughout the prediction horizon, denoted as $N$. In the  experiments performed, $N$ is set to $40$ and the controller frequency is $\unit[10]{Hz}$, which means that the planning horizon is 4 seconds and with a time step of $\unit[0.1]{s} = \tau$. This allows us to express the position of the obstacle at time step $n$ in the future as $\hat{o}_{k+n\mid k} = b^{\mathcal{G}}_{k} + v^{\mathcal{G}}  \tau n$ from time step $k$.

\subsection{Clustering}
When pedestrians move in groups, it is not necessary to treat each person as a separate obstacle. Instead, a reasonable avoidance policy is to avoid the entire group. This approach reduces the number of obstacles that the NMPC needs to consider, making it easier to actively avoid a larger number of individual pedestrians without introducing local minima or large computation requirements.

To group pedestrians hierarchical clustering is used. All obstacles that are closer than two safety marginal '$2s$' at the beginning and the end of the prediction horizon '$N$' are grouped into one cluster '$c$'. The following equation is used to determine if an obstacle is within a cluster:
\begin{align}
\begin{split}
    o_{i} \in~c :  
    \inference{\exists~o_{c} \in~c}{\sqrt{o_{i}^{2} - o_{c}^{2}} < 2s
    \wedge~\sqrt{o_{i,N}^{2} - o_{c,N}^{2}} < 2s}
\end{split}
\label{eq:setes}
\end{align}
Any $\hat{o}$'s that are not sorted into any $c$ are added to the set of clusters $C$ as a cluster with only one object in it. 
If two $c$ are closer together than $2s$ at the current time and at $N$ are they merged to form one bigger $c$. 

Figure~\ref {fig:cluster} shows four different cases with respect to clustering. When pedestrians stand more than $2s$ apart, they form two separate clusters (as shown in Figure~\ref{fig:cluster:a}). If they are slightly closer, they will form one cluster (as shown in Figure~\ref{fig:cluster:b}). If two pedestrians are moving in the same direction, they form one cluster, while they form separate clusters if they are moving in different directions (as shown in Figure~\ref{fig:cluster:c} and~\ref{fig:cluster:d}). 
\input{clustering.tex} 

\subsection{Ellipse collision boxes}
The Nonlinear Model Predictive Control (NMPC) requires a geometric formula to describe the unsafe space created by obstacles in the cluster. If $\hat{o}$ is randomly placed in $c$, a circular enclosure can become unnecessarily large if the $\hat{o}$ are distributed on a line. On the other hand, a minimum spanning ellipse can enclose $\forall \hat{o} \in c$ with a smaller area. The algorithm by Gr{\ae}tner and Sh{\"o}nherr~\cite{gartner1997smallest} is used, along with their implementation~\cite{gartner1998smallestImpl}. The result is a general ellipse equation that describes the edge of the ellipse:
\begin{align}
    e_1x^2 + e_2y^2 + e_3xy + e_4x + e_5y + e_6 = 0
\end{align}
This equation describes the minimum spanning ellipse that includes all obstacles in a cluster, but it does not account for any safety marginal.
Therefore, the equation is transformed into the canonical form: 
\begingroup
\allowdisplaybreaks
\begin{align}
\Gamma &= e_1 e_5^2 + e_3 e_4^2 - e_2 e_4 e_5 + (e_2^2 - 4 e_1 e_3) e_6\\
\gamma &= (e_1 - e_3)^2 + e_2^2\\
E_{major} &=  \frac{-\sqrt{2\Gamma((e_1 + e_3) + \sqrt{\gamma})}}{e_2^2 - 4 e_1 e_3} \\ 
E_{minor} &=  \frac{-\sqrt{2\Gamma((e_1 + e_3) - \sqrt{\gamma})}}{e_2^2 - 4 e_1 e_3} \\
E_x &= \frac{2 e_3 e_4 - e_2 e_5}{e_2^2 - 4 e_1 e_3} \\
E_y &= \frac{2 e_1 e_5 - e_2 e_4}{e_2^2 - 4 e_1 e_3} \\
E_{\theta} &= 0.5\atantwo (\frac{e_2}{e_3 - e_1})
\end{align}
\endgroup
where $s$ is added to the semi-major $E_{major}$ and semi-minor $E_{minor}$ axes. The set $\bm{E}$ of ellipse equations $E$ (one for each $c \in C$) with the added safety marginal describes the set of unsafe spaces.

With the constant speed assumption, it is assumed that the unsafe space will maintain its form and the average speed of the included objects is used to determine the ellipse's speed. Thus, the position at time step $n$ of the unsafe space can be described as follows:
\begin{align}
E_{x,k+n\mid k} &= E_{x} + \bar{c}_{vx} \tau n\\
E_{y,k+n\mid k} &= E_{y} + \bar{c}_{vy} \tau n
\end{align}
with $\tau$ denoting the sampling time. Following the previous steps, we can form an inequality statement for any point $p$ inside the unsafe set as:

\begin{align}
\begin{split}
   E_{major}^2 E_{minor}^2 - E_{major}^2 (\sin E_{\theta})^2 - E_{minor}(\sin E_{\theta})^2
   (p_{x} - E_{x,k+n\mid k})^2 \\+ 2 (E_{minor}^2 - E_{major}^2) \sin (E_{\theta}) \cos (E_{\theta}) (p_{x} - E_{x,k+n\mid k})(p_{y} - E_{y,k+n\mid k}) \\
    + (E_{major}^2 (\cos E_{\theta})^2 + E_{minor}^2 (\sin E_{\theta})^2) (p_{y} - E_{y,k+n\mid k})^2 \leq 0
\end{split}
\label{eq:colition}
\end{align}

Which can then be used to form obstacle avoidance constraints in the nonlinear MPC described in the subsequent Section. For the easiness of the  notations, let us summarize the full parameterization of the generalized ellipse as $\overline{\bm{E}}$.

\subsection{Model Predictive Controller}\label{sec:mpc}
Collision avoidance maneuvers in the proposed DTAA are generated by a high-performance NMPC framework~\cite{lindqvist2020dynamic, lindqvist2021reactive} that is implemented in the Optimization Engine~\cite{sopasakis2020open} - capable of constrained nonlinear optimization with real-time applications making it a perfect fits for the DTAA. This controller closely follows our previous work~\cite{karlsson2022ensuring} on legged robot avoidance but we summarize the framework for completeness and highlight the novelties in this section.

The controller considers a simple kinematic model of the system with the states $x = [x,y, v_x, v_y, \theta]$ as the position, velocity, and heading states, and control inputs $u = [u_{vx}^\mathcal{B}, u_{vy}^\mathcal{B}, w]$, as the control velocities in the body frame $\mathcal{B}$, as well as a heading rate command. Next, the cost function defines the control objective of the NMPC. In this case, we aim to track the desired state trajectory $\psi$, while minimizing actuation and delivering smooth control signals. The cost function $J(\bm{x, u},u_{n-1})$ follows the common quadratic costs on states, control inputs, and an added control input rate cost along the prediction horizon $N$. 

To enable collision avoidance constraints OpEn applies a penalty method~\cite{hermans2021penalty}. To fit the ellipsoid bounding boxes into the OpEn framework we use the $max(a,0) = [a]_{+}$ function. With this simple re-write, we can take the inequality in \eqref{eq:colition} and turn it into an equality constraint as:
\begin{equation}
    C_\mathrm{obs}:= [E_{major}^2 E_{minor}^2 \ldots]_{+} = 0
\end{equation}
such that the constraint is only non-zero if the robot position states are inside the ellipse at the current or any predicted time step. This formulation of general ellipse bounding boxes, as opposed to circles or polygons, allows us to capture better the unsafe space containing multiple obstacles, as a novelty to the previous works~\cite{karlsson2022ensuring, lindqvist2021reactive}. As the DTAA uses a camera to track objects, we must also ensure that obstacles stay inside the camera's field of view. We thus form a similar constraint on tracking the obstacle cluster:
\begin{equation}\label{eq:track_constraint}
       C_\mathrm{track}(\theta, \theta_\mathrm{ref}, \beta) = [-\cos(\theta_{ref} - \theta) + 1 - \beta]_{+} = 0
\end{equation}

Where $\beta$ represents a value smaller than half the camera field of view to have a margin for not losing track of the object, and $\theta_{ref}$ denotes heading references looking toward the predicted position of the center of the obstacle cluster from the predicted position of the robot. Finally, we also impose hard bounds on the control inputs as $u_\mathrm{min}\leq u_{k+n|k} \leq u_\mathrm{max}.$ This results in the following NMPC problem:

 \begin{subequations}\label{eq:nmpc}
\begin{align}
    \operatorname*{Minimize}_{
        \bm{u}_k, \bm{x}_k 
    } \,
    & J(\bm{x}_{k}, \bm{u}_{k}, u_{k-1\mid k}) \notag
    \\
    \text{s. t.:}\,& 
    x_{k+n+1\mid k} = \zeta(x_{k+n\mid k}, u_{k+n\mid k}),\notag
     \\ & n=0,\ldots, N, \notag
    \\
    &u_{\min} \leq u_{k+n\mid k} \leq u_{\max},\, n=0,\ldots, N, \notag
    \\
    &C_{\mathrm{obs}}(\overline{\bm{E}}_{n,j}, p_{k+n \mid k}) = 0, \notag n=0,\ldots, N, \notag\\
    & j = 1, \ldots N_{obs} \notag \\
    &C_{\mathrm{track}}(\theta_\mathrm{k+n|k}, \theta_\mathrm{ref,n}, \beta) = 0,\notag\\
     &n=0,\ldots, N, \notag\\
    &x_{k\mid k} {}={} \hat{x}_{k}. \tag{\ref{eq:nmpc}}
\end{align}
\end{subequations}
with $N_{obs}$ denoting the number of ellipse-constraints. Through a \textit{single shooting approach}, this NMPC problem fits into the framework of the Optimization Engine~\cite{sopasakis2020open}, which we use to solve for trajectories that satisfy the obstacle and tracking constraints while minimally deviating from the tracking trajectory $\psi$.



\subsubsection{Multiple objects}
Our previous work~\cite{karlsson2022ensuring} focused on avoiding one obstacle, whereas this work extends to multiple unsafe sets. However, the constraint in \eqref{eq:track_constraint} can only be applied to keep one specific cluster in the field of view. Many obstacle clusters can also introduce computation problems. As such, we limit our evaluation to $N_{obs} = 2$. Still, we need to prioritize the generated clusters / unsafe sets for tracking, in case more than 2 are in proximity. The prioritization defines the most dangerous cluster as the one closest to $\hat{x}_{n}$ within the prediction horizon as:
\begin{align}
\begin{split}
\Delta_{min} = min\left(\sqrt{(\hat{x}_{x,n} - E_{x,n})^2 + (\hat{x}_{y,n} - E_{y,n})^2}\right)
\end{split}
\end{align}
If two $E$ have the same $\Delta_{min}$ (distance between Spot and the unsafe set), the $E$ that first (lowest $n$) causes the closest $\Delta_{min}$, is considered to have a lower distance. Thus, the $E$ with the smallest $\Delta_{min}$ is prioritized for tracking, and the NMPC attempts to keep the most dangerous cluster within the camera field of view. However, if any cluster $E$ has $\Delta_{min}> \unit[8]{m}$, it is ignored, and the NMPC heading is set to be aligned to the tracked path $\psi$.


\subsubsection{Prediction horizon}
The number of time steps $k$ the NMPC predicts greatly impacts the convergence time for $x_{ref}$.
The controller running in \unit[10]{hz} creates a natural time limit for computation at \unit[100]{ms} to have control actions before it is time to calculate the next. The NMPC has a watchdog on \unit[100]{ms} where if there is no calculated $u$ that satisfies~\eqref {eq:nmpc} it will terminate and return the previous calculated control value.

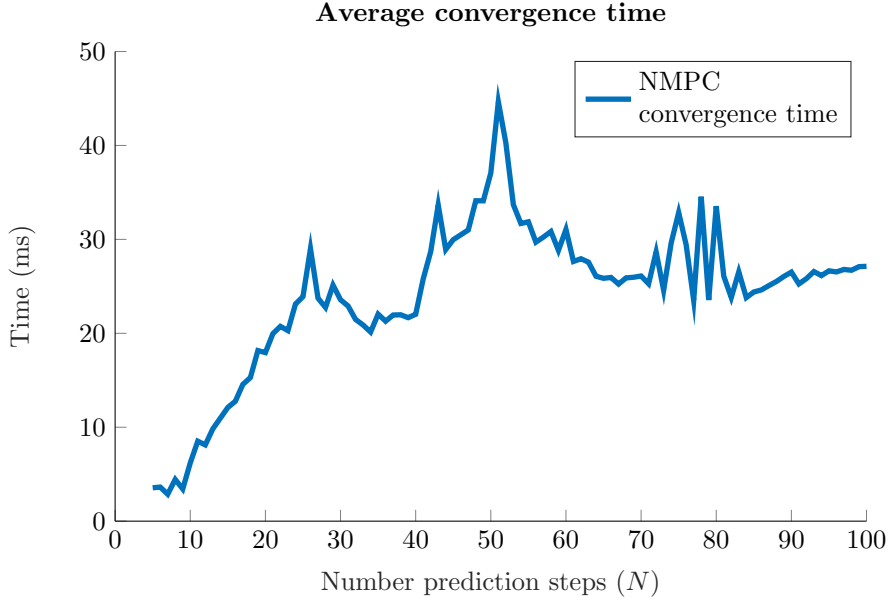
\begin{figure}
    \setlength\fwidth{0.95\columnwidth}
%
%
\definecolor{mycolor1}{rgb}{0.00000,0.44700,0.74100}%
\begin{tikzpicture}

\begin{axis}[%
width=0.8\fwidth,
height=0.5\fwidth,
at={(0.9\fwidth,0.9\fwidth)},
scale only axis,
xmin=0,
xmax=100,
xlabel style={font=\color{white!15!black}},
xlabel={Number prediction steps ($N$)},
ymin=0,
ymax=50,
ylabel style={font=\color{white!15!black}},
ylabel={Time (\unit[]{ms})},
axis background/.style={fill=white},
title style={font=\bfseries},
title={Average convergence time},
axis x line*=bottom,
axis y line*=left,
legend style={legend cell align=left, align=left, draw=white!15!black}
]
\addplot [color=mycolor1, line width=2.0pt]
  table[row sep=crcr]{%
5	3.55496099062579\\
6	3.62602387123263\\
7	2.86871813042246\\
8	4.44015937546043\\
9	3.42575254748104\\
10	6.19981807700023\\
11	8.51166869294317\\
12	8.1188921633881\\
13	9.82271472362378\\
14	10.9612916533731\\
15	12.1045561061005\\
16	12.763853767093\\
17	14.5610513202208\\
18	15.2853667823354\\
19	18.1587165220779\\
20	17.947387786393\\
21	19.9659092904614\\
22	20.735871790431\\
23	20.2877316974438\\
24	23.112286746969\\
25	23.9031794636689\\
26	29.0072917095928\\
27	23.7415006318013\\
28	22.7521602215601\\
29	25.129729325474\\
30	23.570045463497\\
31	22.9107200258127\\
32	21.4759824184839\\
33	20.885477213913\\
34	20.1360698962467\\
35	22.0414614239792\\
36	21.286552994661\\
37	21.9459612747668\\
38	21.9708633817764\\
39	21.6665757458314\\
40	22.0237221321941\\
41	25.7850934422485\\
42	28.7009654580513\\
43	33.6603915038518\\
44	28.9199903740902\\
45	29.9691430046202\\
46	30.4836729373006\\
47	31.0002299594852\\
48	34.1120170530048\\
49	34.1044418097133\\
50	37.0716001175369\\
51	44.6481776077254\\
52	40.2846394483739\\
53	33.6906552607634\\
54	31.6935378638892\\
55	31.8673237145564\\
56	29.7020896376453\\
57	30.2439536817307\\
58	30.8397695014445\\
59	28.8313153897806\\
60	31.0515865928604\\
61	27.6499426578203\\
62	27.9375068225918\\
63	27.5728007504823\\
64	26.073785938374\\
65	25.8552808772391\\
66	25.9441678447386\\
67	25.2393623841832\\
68	25.9015584988432\\
69	25.9655553151748\\
70	26.0989982722653\\
71	25.2754253642159\\
72	28.6627769550674\\
73	24.5790346906523\\
74	29.6061184481994\\
75	32.8537604297082\\
76	29.4070102372505\\
77	23.5101232766843\\
78	34.5473125891974\\
79	23.541920979668\\
80	33.5390485461231\\
81	26.0913989808649\\
82	23.8110652933444\\
83	26.5427341969051\\
84	23.7992285729564\\
85	24.3992481219352\\
86	24.6162593637491\\
87	25.0758340571711\\
88	25.5140388090613\\
89	26.0523585703599\\
90	26.5157499169688\\
91	25.2578447616664\\
92	25.8211217592727\\
93	26.573245309741\\
94	26.1416218520922\\
95	26.651477811489\\
96	26.5350741654123\\
97	26.7998152817846\\
98	26.7112181274399\\
99	27.0941658231896\\
100	27.1320316949415\\
};
\addlegendentry{NMPC \\convergence time}

\end{axis}
\end{tikzpicture}%
    \caption{The average of \unit[]{ms} for the NMPC to converge motion actions for different prediction horizons ($N$). These measurements are from an offline analysis of a priori recorded data.}
    \label{plot:avg}
\end{figure}

\begin{figure}
    \setlength\fwidth{0.95\columnwidth}
%
%
\definecolor{mycolor1}{rgb}{0.00000,0.44700,0.74100}%
\begin{tikzpicture}

\begin{axis}[%
width=0.8\fwidth,
height=0.6\fwidth,
at={(0.9\fwidth,0.9\fwidth)},
scale only axis,
xmin=0,
xmax=100,
xlabel style={font=\color{white!15!black}},
xlabel={Number prediction steps (N)},
ymode=log,
ymin=0,
ymax=10000,
yminorticks=true,
ylabel style={font=\color{white!15!black}},
ylabel={Number of timeouts},
axis background/.style={fill=white},
title style={font=\bfseries},
title={Timeout annalyce},
axis x line*=bottom,
axis y line*=left,
legend style={legend cell align=left, align=left, draw=white!15!black}
]
\addplot [color=mycolor1, line width=2.0pt]
  table[row sep=crcr]{%
5	0.1\\
6	0.1\\
7	0.1\\
8	0.1\\
9	0.1\\
10	2\\
11	2\\
12	2\\
13	1\\
14	0.1\\
15	3\\
16	0.1\\
17	2\\
18	4\\
19	3\\
20	1\\
21	5\\
22	4\\
23	5\\
24	5\\
25	4\\
26	63\\
27	5\\
28	2\\
29	0.1\\
30	2\\
31	4\\
32	2\\
33	1\\
34	8\\
35	7\\
36	5\\
37	2\\
38	8\\
39	4\\
40	7\\
41	145\\
42	16\\
43	1570\\
44	8\\
45	7\\
46	8\\
47	3\\
48	5\\
49	7\\
50	15\\
51	2522\\
52	618\\
53	11\\
54	11\\
55	9\\
56	8\\
57	3\\
58	8\\
59	1\\
60	9\\
61	10\\
62	9\\
63	5\\
64	9\\
65	13\\
66	6\\
67	8\\
68	19\\
69	14\\
70	17\\
71	8\\
72	26\\
73	16\\
74	13\\
75	20\\
76	2189\\
77	549\\
78	1177\\
79	47\\
80	316\\
81	56\\
82	48\\
83	32\\
84	44\\
85	47\\
86	22\\
87	40\\
88	25\\
89	43\\
90	47\\
91	24\\
92	41\\
93	43\\
94	38\\
95	46\\
96	52\\
97	45\\
98	31\\
99	44\\
100	33\\
};
\addlegendentry{NMPC number of timeouts}

\end{axis}
\end{tikzpicture}%
    \caption{The number of times the NMPC took more than \unit[100]{ms} to converge to a solution for different prediction horizons ($N$). These measurements are from an offline analysis of prior recorded data.}
    \label{plot:timeout}
\end{figure}
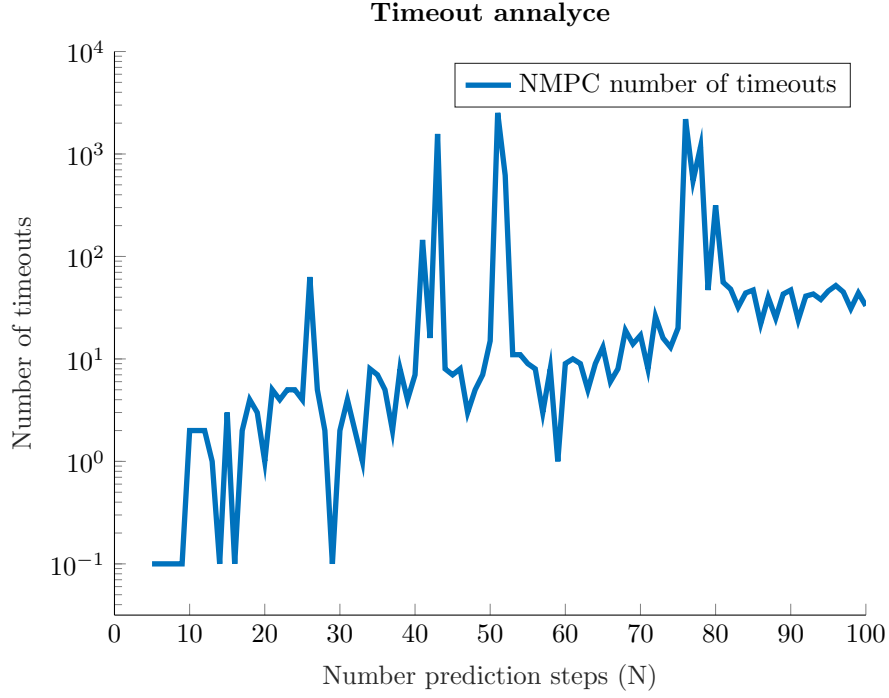

Larger prediction horizons create avoidance maneuvers thunder into the future that can allow for earlier avoidance of objects. But longer horizons also produce larger uncertainty, so there is a limit on how far it is reasonable to predict in the future. By offline running the NMPC with bagged data as input, computation times are recorded for different $N$ between $5-100$. With different $N$ both the NMPC's planning steps into the future and the obstacle future position prediction steps are altered.
The average calculation time can be seen in Figure~\ref{plot:avg}, while in Figure~\ref{plot:timeout} the number of times that the NMPS exceeded the maximum allowed computation time of \unit[100]{ms} is depicted. Because of the varied horizons utilized, finding a valid path is sometimes harder than other times, as it could be observed from the spikes in Figure~\ref{plot:timeout}. One explanation for the spikes in Figure~\ref{plot:timeout} could be mainly due to the fact that the predicted position of $c$ at time $N$ is close to the desired end position with that particular horizon. The average converging time has a logarithmic trend, where $N$ increases, suggesting that $N=100$ is possible. However, the likelihood of timeouts also increases as $N$ increases, which is potentially dangerous and can lead to bad solutions. At the same time, the accuracy of the linear assumption droops with higher $N$, with this information, and thus is why $N=40$ was decided to be the $N$ value for the experiments.


\subsection{Robotic mission components, Path planning} 
In this article an evolution of D$^{*}$-lite has been utilized that it is an evolution of the A$^{*}$ path planning algorithm called D$^{*}_{+}$~\cite{karlsson2022d+}. D$^{*}_{+}$ is a path planner that considers unknown areas and areas close to obstacles, such as walls, as a risk when it plans a path. The risks are considered as a traversal cost $\xi$ in addition to the standard distance costs/heuristics. D$^{*}_{+}$ is planning the shortest safe path $\Psi$ from the current location $\hat{x}$, according to the sum of traversal costs in the path $\sum{\forall \xi \in \Psi}$, to the waypoint $WP$.
The occupancy grid map $\mathbf{M}$ used for planning is provided, in this case, by the traversability mapper~\cite{bayesian2018shan}.
Because $\mathbf{M}$ is cells in a grid forming a map, D$^{*}_{+}$ is planning $\Psi$ as the sequence of cells for the robot to visit, in order to move from $\hat{x}$ to $WP$ in the most optimal (short and safe) manner.
The traversal cost $\xi$, for each cell, is calculated as follows:
\begin{align}
    \xi_r &= \begin{cases}
        \frac{\xi_u}{d + 1} & \text{if $d < r$} \\
        0 & \text{else} \\
    \end{cases}\\
    \xi &= \begin{cases}
        \xi_r + \xi_d& \text{if the cell is free} \\
        \xi_u + \xi_r + \xi_d& \text{if the cell status is unknown}
    \end{cases}
\end{align}
where $d$ is the distance from the cell to the closest occupied cell and $r$ is the radius in which risk is considered.
$\xi_u$ is the designed cost for unknown cell, and $\xi_d$ is the cost for the travel distance.
The risk cost is noted as $\xi_r$ and is a part of the total cost $\xi$ to traverse through that cell.

\section{Experiments}

\begin{figure}
    \centering

\begin{subfigure}[t]{0.9\columnwidth}
   \centering
   \includegraphics[width=0.9\linewidth]{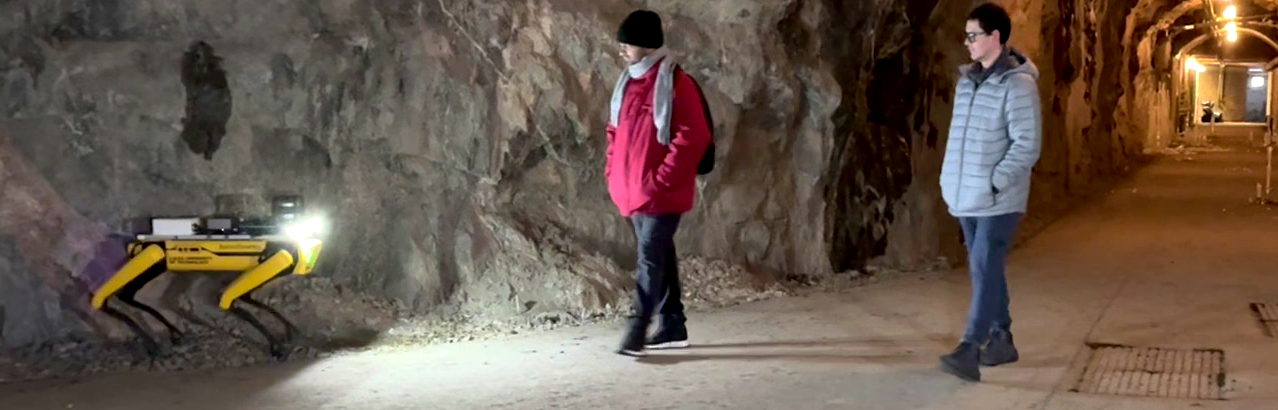}
   \label{fig:snap:cluster}
 \end{subfigure}
 \hfill
\begin{subfigure}[t]{1.0\columnwidth}
   \centering

    \resizebox{0.8\columnwidth}{!}{%
\begin{tikzpicture}
    \node at (0,0) {\includegraphics[width=0.9\columnwidth]{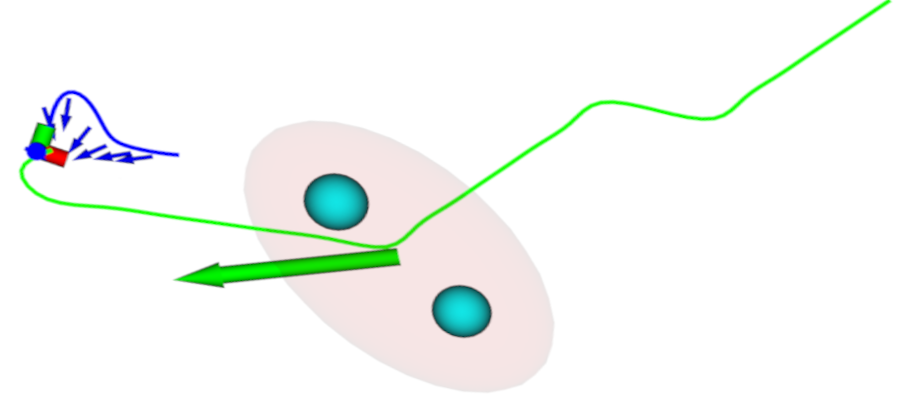}};

     \node[] at (2.0, 0.9) (dsp) {$\Psi$};
     \node[] at (-5.9, 0.7) (s) {$\hat{x}$};
     \node[align=left] at (-3.8,  1.6) (motion) {$u$ with heading};
     \node[align=center] at (0.0, -0.3) (obs) {O};
     \node[align=center] at (-0.2, 0.7) (elips) {$E$};
     \node[align=center] at (-3.9, -0.9) (ev) {$E_v$};
     \draw[-stealth] (obs) -> (-1.0, 0.0);
     \draw[-stealth] (obs) -> (0.1, -0.95);




    
\end{tikzpicture}
}
    \caption{The legends used in this article for data visualization. The axes represent the Spot position. The green line is the path from D$^{*}_{+}$, while the blue line is the predicted motion of Spot along with arrows indicating its heading. The teal circles are the detected pedestrians, which are enclosed by the bounding ellipse with added safety marginal (unsafe space). Finally, the green arrow denotes the velocity of the unsafe space.}
    \label{fig:legen}
\end{subfigure}
    \caption{Spot is meeting 2 pedestrians while moving around a bend. To avoid collisions spot moving close to the wall to let the pedestrians pass. As well as planning to adjust the heading as the pedestrians pass to keep them in the field of view.}
    \label{fig:snap:meeting}
\end{figure}

Given the challenge of describing dynamic motion through text and figures, we recommend watching the video at \url{https://youtu.be/VJRxoYuXA0I} for a comprehensive understanding. Data visualization is extensively used throughout the articles, while figure~\ref{fig:legen} serves as a legend. 

\subsection{Lab experiments}
\begin{figure}
    \setlength\fwidth{0.95\columnwidth}
    \input{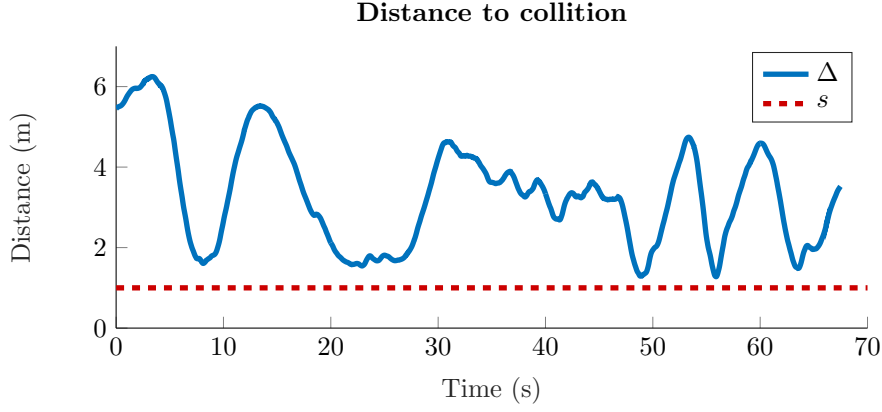}
    \caption{Distance between spot and an obstacle during a lab experiment.}
    \label{plot:dist}
\end{figure}

\begin{figure}
    \setlength\fwidth{0.95\columnwidth}
    \input{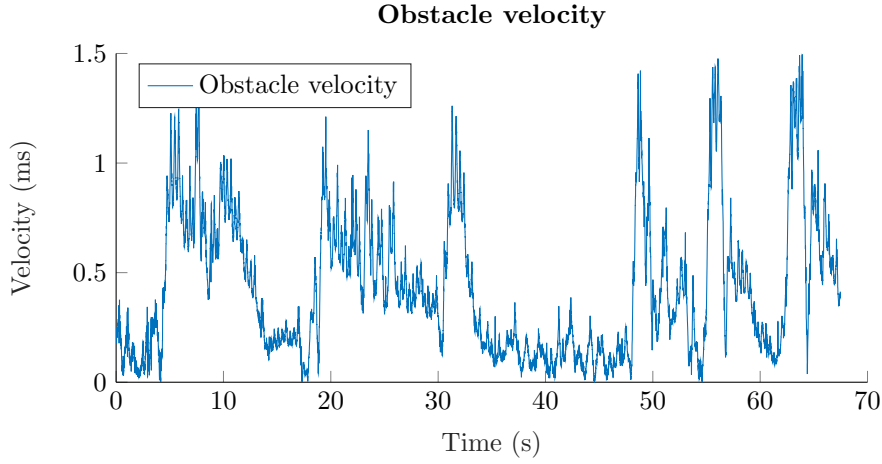}
    \caption{Velocity of the obstacle in the same experiment as in Figure~\ref{plot:dist}.}
    \label{plot:vel}
\end{figure}

During the lab testing, DTAA was evaluated in various scenarios involving a pedestrian moving towards Spot, and Spot moving past a stationary pedestrian, as well as in meeting situations. Additionally, the tracking ability of DTAA, with a heading angle, was tested by having a pedestrian running across Spot's field of view (Please refer to Figure~\ref{fig:snap:tracking} for a visual representation of the tracking test). The blue arrows show the heading that the NMPC is planning in order to keep the pedestrian in the field of view.
\input{tracking.tex}
During the experiment, we measured the distance between Spot and the pedestrian ($\Delta$) using a motion capture system that provided a ground-truth distance. As shown in Figure~\ref{plot:dist}, the safety marginal $s = \unit[1]{m}$ between Spot and the pedestrian has never been violated $\forall k+n, (\Delta > s)$. Even though the pedestrian approached Spot in approximately $\unit[1.4]{ms}$, as shown in Figure~\ref{plot:vel}.

\subsection{Corridor experiments}
The lab experiments were conducted again, but this time in a corridor, with three pedestrians entering the robots' space. This provided a different environment to observe. The clustering technique was observed during active motion, as shown in Figure~\ref{fig:snap:cluster}.
\input{clustering.tex}
During the tests, Spot was able to avoid collisions and maneuver around the pedestrians, as shown in Figure~\ref{fig:snap:around} where it rounded (red line) a stationary unsafe space. The avoidance maneuver (blue line) is a minimal divergence from $\Psi$, while maintaining $\Delta > s$ and for Spot to get to the desired position.
\input{arounde.tex}

\begin{figure}
    \setlength\fwidth{0.9\columnwidth}
    \input{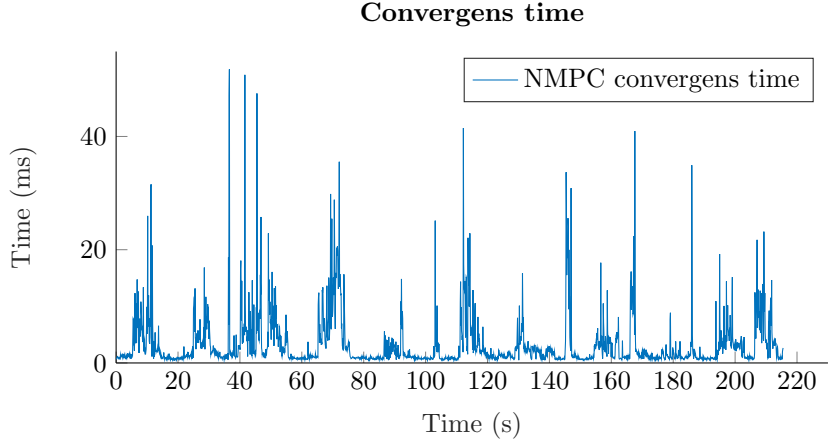}
    \caption{The NMPC time of convergence throughout the entire corridor experiment.}
    \label{plot:convergens}
\end{figure}

Figure~\ref{plot:convergens} shows the time taken by the NMPC to create a suitable motion plan to avoid obstacles. It is evident that more time is required when the NMPC is challenged with avoidance. However, in this experiment, the NMPC always converged within the $\unit[100]{ms}$ time limit and never failed.

\subsection{Outdoor experiment}
To assess DTAA's robustness in varied environments, an outdoor experiment was conducted with Spot traversing an extended $\Psi$ for enhanced realism. Spot encountered two approaching pedestrians with ample spacing between them for Spot to navigate through. As depicted in Figure~\ref{fig:snap:oute}, Spot effectively initiated an avoidance maneuver by deviating from $\Psi$ well before a collision was imminent, showcasing the system's predictive capabilities. The upper/left unsafe space is prioritized to have inside the field of view, evidenced by the blue arrows indicating Spots' future viewing angle.
\input{regn.tex}

\subsection{Subterranean experiments}
DTAA also underwent testing in a mine-like subterranean environment, which presents unique challenges, primarily limited visibility due to darkness and an unknown and unstructured harsh terrain to transverse. In this case, most experiments were conducted with supplementary external lighting and Spot-mounted lights to overcome this difficulty.

Throughout the DTAA experiments in this environment, only two situations occurred where the safety margin was violated. The first case occurred when DTAA avoided an obstacle, and someone approached from behind, outside the camera's field of view. The second instance was observed when the approach speed was exceptionally high, often accompanied by sudden acceleration from relative proximity.

\subsubsection{Stationary Spot}
In this test scenario, illustrated in Figure~\ref{fig:snap:hold3}, three pedestrians traverse Spot's position, making Spot execute an avoidance maneuver. During this maneuver, Spot briefly loses sight of all three pedestrians, resulting in two clusters in the third snapshot upon completion of the avoidance. In this sequence is it also observed how the NMPC reacts when a collision is predicted by initiating an avoidance maneuver but still with plenty of marginal space before driving to a collision. After the avoidance maneuver is completed, Spot succesfully returns to its original position.
\input{hold3.tex}

\subsubsection{Moving spot}
Multiple experiments were carried out to assess Spot's waypoint navigation capabilities. In one instance, while moving along $\Psi$, Spot encountered three pedestrians approaching from different angles and speeds, as depicted in Figure~\ref{fig:snap:diffspeed}. Spot effectively avoided a collision by sidestepping and subsequently continuing forward.
\input{diffspeed.tex}

In a different scenario, Spot encountered two groups of two pedestrians each, as shown in Figure~\ref{fig:snap:grup2}. Spot initially attempted to move aside to let the pedestrians pass, but a wall blocked its path, leading to a backup maneuver to avoid a collision. Despite encountering a local minimum and temporarily struggling to find a way out, Spot successfully avoided a collision.
\input{group2.tex}

In another scenario, Spot adeptly moved to the side to allow two pedestrians to pass by, as depicted in Figure~\ref{fig:snap:meeting}. This interaction occurred as Spot navigated a path around a bend, and the pedestrians approached from the opposite direction.

In another experiment at Sadviks test mine in Finland, DTAA has been evaluated while following a $\Psi$ around a corner. A couple of mine workers stood on the other side of the corner. When Spot was about to pass them, one of the workers broke free from the group and crossed $\Psi$ forcing an avoidance maneuver before reattaching the end of $\Psi$. The entire sequence is shown in Figure~\ref{fig:sandvik}

\input{sandvik.tex}
\subsubsection{Traversing a dark tunel}
During this mission, Spot had to traverse a dark tunnel with a long $\Psi$ without any external light source. in this case, the navigation relied solely on its onboard lights to guide it. While in the tunnel, Spot encountered, in a curve, two pedestrians on the inside and had to plan an obstacle free path. A snapshot of this encounter is displayed in Figure~\ref{fig:snap:dark}. Although the pedestrians are hard to see in the image, YOLOv8 was able to detect them. Although the pedestrians were detected and successfully avoided it is also clear that DTAA struggles with the darkness as evidenced by the multiple detected unsafe spaces seen in Figure~\ref{fig:snap:dark}. 
\input{dark.tex}
\subsubsection{Comparison with only LiDAR-based artificial potential field}
\begin{figure}
    \setlength\fwidth{0.95\columnwidth}
    \input{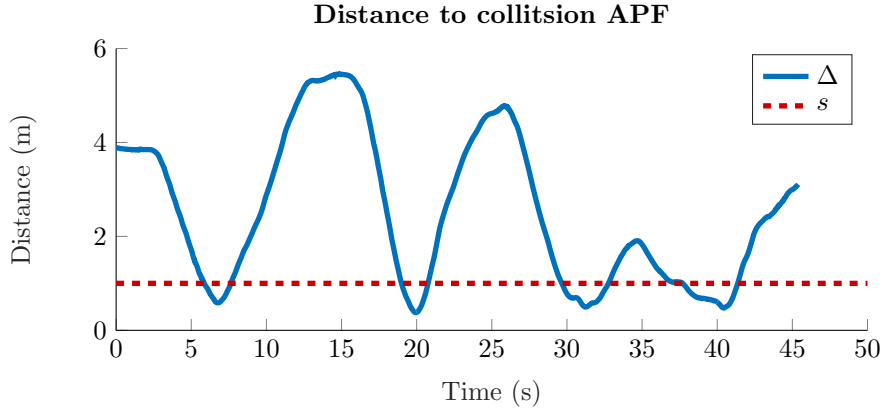}
    \caption{Distance between Spot and the obstacle when using only APF to avoid collisions.}
    \label{plot:apf}
\end{figure}

The Artificial Potential Field~\cite{lindqvist2022adaptive} (APF) responds when the $\unit[1]{m}$ radius of influence is breached, as illustrated in Figure~\ref{plot:apf}. Due to its reactive nature, the smallest distance is inherently shorter than the radius of influence. While a larger radius can prevent safety margin violations, it may result in unnecessary avoidance maneuvers, especially when encountering pedestrians walking alongside Spot. Additionally, a larger radius increases the risk of local minima, a susceptibility already present in APF. Nevertheless, the APF is deemed effective as it ensures collision avoidance in all directions and remains unaffected by darkness.

\subsubsection{Comparison with Spot built-in avoidance}
Spot's collision avoidance operates at the low-level controller after NMPC, utilizing data from its built-in cameras. When provided with velocity commands, Spot executes them while avoiding collisions. With collision safety margin set to $s$ Spot is moving to the side of obstacles, similar to DTAA (see Figure~\ref{fig:internal}). However, this internal solution is not always reliable; although it might not fail outright, the safety margin can be breached (see Figure~\ref{fig:internal:fail}). A notable drawback is in the inability of the framework to handle velocities, rendering it susceptible to moving obstacles.

\input{example.tex}

\section{Conclusions}
This article introduces the Detect, Track, and Avoid Architecture (DTAA), expanding on prior work~\cite{karlsson2022ensuring} by enabling simultaneous avoidance of multiple velocity obstacles. DTAA employs clustering and prioritization based on YOLOv8 object detections for reliable collision avoidance, as demonstrated through extensive testing in various scenarios. A comparison with APF and Spot's internal solution underscored the DTAA's benefits.

Despite its success, DTAA has limitations, such as difficulty in detecting pedestrians outside its field of view and fast-approaching obstacles with rapid acceleration. These limitations stem from constraints rather than method flaws. Potential solutions include mounting additional cameras for full 360$^{\circ}$ vision and running multiple YOLOv8 instances. However, handling high-velocity objects presents a more intricate challenge, requiring improvements to the system for mitigation.

Even when DTAA fails to move around an obstacle and gets stuck in a local minima, it has a safe behavior, since it is better to reverse and get stuck than to collide. To solve this issue, probably a combined solution where one module does both path planning and avoidance is needed. 

\section*{Declarations}
The authors declare that no funds, grants, or other support were received during the preparation of this manuscript.\\
The authors have no relevant financial or non-financial interests to disclose.\\
No ethical concerns apply to this work.\\
The authors affirm that all participants in all of the experiments approve with informed consent to experiments and publication.\\
The work was divided between the authors as follows:
Samuel Nordstr{\"o}m did the main implementation of software where responsible for the execution of the experiments, he also wrote the main part of the article. Bj{\"o}rn Lindquist designed the NMPC and wrote the NMPC section as well as improved the quality of the rest of the article. George Nikolakopoulos supervised the work and has done proofreading of the article.


\bibliography{sn-bibliography}

\end{document}